\documentclass[10pt,twocolumn,letterpaper]{article}

\usepackage[pagenumbers]{cvpr} 

\definecolor{cvprblue}{rgb}{0.21,0.49,0.74}

\usepackage{algorithm}
\usepackage{algpseudocode}
\usepackage{multirow}
\usepackage{tabularx}
\usepackage{subcaption}
\usepackage{enumitem}
\usepackage{pifont}
\newcommand{\cmark}{\ding{51}}%
\newcommand{\xmark}{\ding{55}}%

\usepackage{amsmath,amsfonts,bm}

\def\eqref#1{equation~\ref{#1}}

\def\1{\bm{1}}

\def\vx{{\bm{x}}}

\def\vz{{\bm{z}}}

\DeclareMathAlphabet{\mathsfit}{\encodingdefault}{\sfdefault}{m}{sl}
\SetMathAlphabet{\mathsfit}{bold}{\encodingdefault}{\sfdefault}{bx}{n}

\def\sI{{\mathbb{I}}}

\DeclareMathOperator*{\argmax}{arg\,max}

\usepackage{graphicx}
\usepackage{amsmath}
\usepackage{amssymb}
\usepackage{booktabs}

\usepackage[pagebackref,breaklinks,colorlinks,allcolors=cvprblue]{hyperref}

\usepackage[capitalize]{cleveref}
\crefname{section}{Sec.}{Secs.}
\Crefname{section}{Section}{Sections}
\Crefname{table}{Table}{Tables}
\crefname{table}{Tab.}{Tabs.}

\def\paperID{17} 
\def\confName{CVPR}
\def\confYear{2025}

\title{Pseudo-labelling meets Label Smoothing for Noisy Partial Label Learning}

\author{
Darshana Saravanan\\
{\tt\small darshana.s@research.iiit.ac.in}
\and
Naresh Manwani\\
{\tt\small naresh.manwani@iiit.ac.in}
\and
Vineet Gandhi\\
{\tt\small vgandhi@iiit.ac.in}
\and
\vspace{-0.5 em}
IIIT Hyderabad, India
}

\begin{document}
\maketitle
\begin{abstract}
We motivate weakly supervised learning as an effective learning paradigm for problems where curating perfectly annotated datasets is expensive and may require domain expertise such as fine-grained classification. We focus on Partial Label Learning (PLL), a weakly-supervised learning paradigm where each training instance is paired with a set of candidate labels (partial label), one of which is the true label. Noisy PLL (NPLL) relaxes this constraint by allowing some partial labels to not contain the true label, enhancing the practicality of the problem. Our work centres on NPLL and presents a framework that initially assigns pseudo-labels to images by exploiting the noisy partial labels through a weighted nearest neighbour algorithm. These pseudo-label and image pairs are then used to train a deep neural network classifier with label smoothing. The classifier's features and predictions are subsequently employed to refine and enhance the accuracy of pseudo-labels. We perform thorough experiments on seven datasets and compare against nine NPLL and PLL methods. We achieve state-of-the-art results in all studied settings from the prior literature, obtaining substantial gains in the simulated fine-grained benchmarks. Further, we show the promising generalisation capability of our framework in realistic, fine-grained, crowd-sourced datasets.
\end{abstract}    
\section{Introduction}
\label{sec:intro}

The effectiveness of contemporary deep learning methods heavily relies on the presence of high-quality annotated data. Although this might be available for typical image classification, curating labeled datasets for fine-grained classification problems
is costly, arduous and requires expert knowledge. Weakly supervised learning offers a potential solution to this challenge by learning from partially labelled or noisily labelled examples. It has been widely studied in different forms, including multi-label learning~\cite{zhang2013review}, semi-supervised learning~\cite{sohn2020fixmatch}, noisy label learning~\cite{li2020dividemix} etc. Our paper focuses on a weakly supervised setting called Partial Label Learning (PLL), where each training instance is paired with a set of candidate labels (partial label), out of which one is the true label. A fundamental limitation of PLL is the assumption that the correct label is always included in the partial label. To overcome this limitation and further the practicality, Noisy PLL (NPLL), also referred to as Unreliable Partial Label Learning (UPLL)~\cite{qiao2023fredis} was proposed. NPLL allows some partial labels to not contain the correct label. In fine-grained classification, annotators may provide multiple possible labels (e.g., multiple bird species that look similar), making NPLL techniques useful for learning from such ambiguous annotations~\cite{briggs2012rank}.



 Prior works~\cite{wang2022pico+,qiao2023fredis,Shi2023Unreliable,lian2023irnet,xu2023alim} on NPLL have predominantly pursued a label disambiguation approach by maintaining and updating a distribution over label probabilities. To tackle noisy samples (samples whose correct label is not in the partial label), current methods apply a detection cum mitigation technique~\cite{wang2022pico+,lian2023irnet,Shi2023Unreliable} which are known to cause error propagation due to the unavoidable detection errors~\cite{xu2023alim}. Further, most state-of-the-art (SOTA) methods~\cite{wang2022pico+,lian2023irnet,xu2023alim} require a two-stage training. The first stage involves warm-up training of the classifier, typically using a PLL algorithm which is followed by their proposed NPLL strategy. Determining the optimal duration for warm-up poses a challenge, since a too long or short warm-up period can negatively impact the performance~\cite{xu2023alim}. Some methods also need a clean validation set to decide the warm-up duration~\cite{Shi2023Unreliable,lian2023irnet}, which may not always be available. 


In this paper, we propose a novel iterative pseudo-labelling based framework for NPLL called PALS, combining \textbf{P}suedo-labelling \textbf{A}nd \textbf{L}abel \textbf{S}moothing. Unlike prior works that use the probability distribution over class labels for each image as the supervision for classifier training, PALS utilizes a pseudo-labelling strategy involving the weighted nearest neighbour algorithm, that assigns a single pseudo-label to each image. PALS builds upon the early efforts of~\cite{hullermeier2005learning}, suggesting that methods featuring a strong inductive bias, such as K-Nearest Neighbors (KNN), can effectively exploit partial label information for improved disambiguation. PALS then uses a selected set of reliable image-label pairs to train the classifier. Our choice of using pseudo-labels instead of soft-label distributions as the supervision, allows us to leverage label smoothing~\cite{szegedy2016rethinking}. This imparts robustness towards the potential errors during the pseudo-labelling stage, specifically in the high noise scenario. Throughout training, we demonstrate that pseudo-labelling enhances the feature representation backbone. This, in turn, enhances the accuracy of pseudo-labelling, creating a positive feedback loop.

In a nutshell, our framework has a single trainable component (e.g. a ResNet-18~\cite{he2016deep} model). It eliminates multi-branch networks~\cite{wang2022pico+} and operates seamlessly without warm-up, resulting in faster training time and lower memory requirements. We perform comprehensive experiments on seven datasets in widely different settings (varying noise rates and number of candidate labels). PALS outperforms the SOTA methods by a significant margin. In contrast to previous methods, PALS maintains its performance largely even as the noise rate increases. Performance improvements, compared to other methods, become more pronounced with increasing numbers of classes and the proportion of noise in partial labels. 

In fine-grained classification, we evaluate PALS in both simulated and real-world crowd-sourced benchmarks. PALS achieves significant improvements in CUB-200, outperforming all other methods by over 5-16pp.  Moving beyond the simulated PLL and NPLL benchmarks as in prior literature, we also show that PALS obtains consistent results on realistic, fine-grained, crowd-sourced datasets~\cite{schmarje2022one}.
In line with prior works, PALS also benefits from standard regularization techniques such as Mix-up and Consistency regularization. However, PALS achieves better performance regardless of the presence or absence of regularization techniques. In summary, our work makes the following contributions:

\begin{itemize}[itemsep=0em]
    \item We propose PALS, a novel iterative pseudo-labelling based framework for NPLL, which involves KNN based pseudo-labelling and label smoothing. To our knowledge, we are the first to showcase the effectiveness of label smoothing for NPLL.
    \item We show notable gains over nine SOTA PLL and NPLL methods in the simulated benchmarks from prior literature. Further, we are the first to demonstrate the effectiveness of NPLL for fine-grained crowd-sourced datasets, with PALS consistently performing well across different datasets and annotator configurations.
    \item We present thorough ablation studies and quantitative experiments to support our design choices and demonstrate the efficacy of our approach. 
\end{itemize}

\section{Related Work}
\label{sec:related_work}

\subsection{Traditional Partial Label Learning}

\textit{Identification-based strategies} (IBS) treat the ground truth as a latent variable and progressively refine the confidence of each candidate label during training. \cite{Jin2002learning} use a maximum likelihood criterion to learn the parameters of a classifier that maximizes the latent label distribution, estimated from the classifier predictions in an EM fashion. \cite{yu2016maximum} propose a maximum margin formulation for PLL, which maximizes the margin between the ground-truth label and all other labels.

\textit{Average-based strategies} (ABS) treat all candidate labels equally while distinguishing between the candidate and non-candidate labels. Early work by \cite{hullermeier2005learning} extends the K-Nearest neighbours (KNN) algorithm to tackle PLL by employing majority voting among the candidate labels of neighbours. \cite{zhang2015solving} also utilize KNN to classify any unseen instance based on minimum error reconstruction from its nearest neighbours. \cite{cour2009learning} maximize the average output of candidate labels and minimize the output of non-candidate labels in parametric models. PALS takes a cue from the KNN-based ABS approach~\cite{hullermeier2005learning} and uses a variation for the pseudo-labelling step. 

\subsection{Deep Partial Label Learning}
With the advent of Deep Learning, a variety of identification-based strategies that employ a neural network backbone have been proposed. \cite{yao2020deep} temporally ensemble the predictions at different epochs to disambiguate the partial label. \cite{lv2020progressive} use self-training to update the model and progressively identify the true label. \cite{feng2020provably} formulate the generation process of partial labels and develop two classifiers: one, risk-consistent and the other, classifier-consistent. \cite{wen2021leveraged} propose a family of loss functions that generalize the different loss functions proposed by earlier works \cite{Jin2002learning,cour2009learning,lv2020progressive}. \cite{wang2022pico} present PiCO, a framework that learns discriminative representations by leveraging contrastive learning and prototypical classification. \cite{wu2022revisiting} present a technique that uses consistency regularization on the candidate set and supervised learning on the non-candidate set. \cite{xia2023towards} upgrade PiCO and introduce self-training and prototypical alignment to achieve noteworthy results. However, none of these methods account for the possibility of noise within partial labels, which is the primary focus of our work.

\subsection{Noisy Partial Label Learning}
Earlier works assume that the correct label is always a part of the partial label, which limits the practicality of the problem. Hence, some recent works have diverted attention to NPLL that relaxes this condition, and allows some partial labels not to contain the correct label. \cite{qiao2023fredis} perform disambiguation to move incorrect labels from candidate labels to non-candidate labels and refinement to move correct labels from non-candidate labels to candidate labels. \cite{Shi2023Unreliable} separate the dataset into reliable and unreliable sets and then perform label disambiguation-based training for the former and semi-supervised learning for the latter. \cite{lian2023irnet} iteratively detect and purify the noisy partial labels, thereby reducing the noise level in the dataset. \cite{wang2022pico+} extend PiCO to tackle noisy PLL by performing distance-based clean sample selection and learning a robust classifier by semi-supervised contrastive learning. \cite{xu2023alim} reduce the negative impact of detection errors by carefully aggregating the partial labels and model outputs.

Unlike the aforementioned identification-focused approaches, \cite{lv2023robustness} propose Average Partial Loss (APL), a family of loss functions that achieve promising results for both PLL and NPLL. Moreover, they provide insights into how ABS can enhance IBS when used for warm-up training. Our work builds upon this intuition by alternating between ABS and IBS. Our findings demonstrate that employing K-nearest neighbours (KNN) for ABS and utilizing cross-entropy loss with label smoothing for IBS, yields SOTA performance.

\section{Methodology}
\label{sec:method}

\begin{figure}[t]
    \centering
    \includegraphics[width=1.\columnwidth]{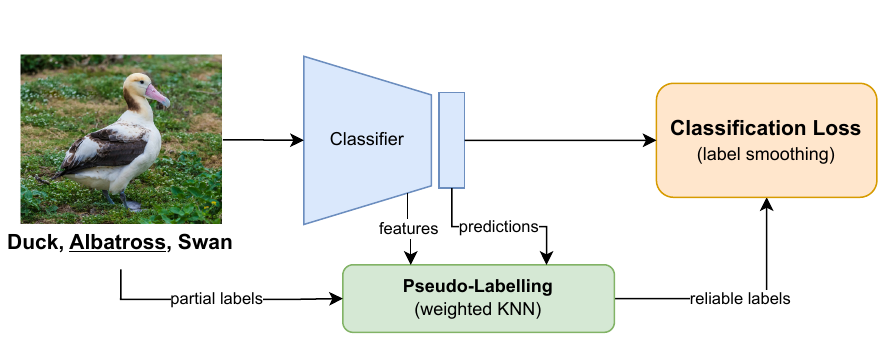}
    \caption{Illustration of PALS. Pseudo-labelling involves the usage  of weighted KNN to obtain reliable image-label pairs. These pairs are used to train a classifier using label smoothing. Finally, the confident top-1 model predictions are used to augment the partial label for the upcoming iteration.}
    \label{fig:PALS}
\end{figure}

Figure~\ref{fig:PALS} depicts the three modules of PALS that are applied sequentially in every iteration: Pseudo-labelling, Noise robust learning and Partial label augmentation. First, we assign pseudo-labels to all images using a weighted KNN. Then, for each class, we select the $m$ most reliable image-label pairs which are used to train the classifier. To be resilient towards the potential noise in the pseudo-labelling stage, we leverage label smoothing. Finally, we augment the partial labels with the confident top-1 predictions of the classifier. The features from the updated classifier are then used for the next stage of pseudo-labelling. We hypothesize that, as training progresses, more number of samples will be assigned the correct label as the pseudo-label, which will help learn an improved classifier. Pseudo-code of PALS is available in Algorithm~\ref{alg:PALS}. We now provide a detailed description of each of the mentioned steps.

\subsection{Problem Setup}

Let $ \mathcal{X}$ be the input space, and ${\mathcal{Y} =\{1,2,\ldots,C\}}$ be the output label space. We consider a training dataset ${\mathcal{D} = \{(\vx_i, Y_i)\}^n_{i=1}}$, where each tuple comprises of an image $\vx_i \in \mathcal{X}$ and a partial label $Y_i \subset \mathcal{Y}$. Similar to a supervised learning setup, PLL aims to learn a classifier that predicts the correct output label. However, the training dataset in PLL contains high ambiguity in the label space, which makes the training more difficult when compared to supervised learning. A basic assumption of PLL is that the correct label $y_i$ is always present in the partial label, i.e., $ \forall i \enspace y_i \in Y_i $. In NPLL, we relax this constraint, and the correct label potentially lies outside the candidate set, i.e., $ \exists i \enspace y_i \notin Y_i$.

\subsection{Algorithm}

\subsubsection{Pseudo-labelling} \label{sec:pseudo_labelling}
For each sample $(\vx_i, Y_i)$, we pass the image $\vx_i$ through the encoder $f(\cdot)$ to obtain the feature representation $\vz_i$. Then, we select its $K$ nearest neighbours from the entire dataset. Let $\mathcal{N}_i$ denote the set of $K$ nearest neighbours of $\vx_i$. We then compute a pseudo-label $\hat{y}_i$ using weighted KNN as follows. 
 
\begin{equation} \label{eq:pseudolabel}
    \hat{y}_i = \argmax_c \sum_{\substack{\vx_k \in \mathcal{N}_i}} \sI[c \in Y_k]\cdot s_{ik} \qquad c \in [C],  
\end{equation}
where $s_{ik}$ is the cosine similarity between $\vz_i$ and $\vz_k$.
We assume that these pseudo-labels are less ambiguous than the partial labels supposing that the samples in the neighbourhood have the same class label. Now, we approximate the class posterior probabilities $\hat{q}_c$ via KNN, inspired by the Noisy Label Learning literature \cite{ortego2021multi}.
\begin{equation} \label{eq:posterior_prob}
     \hat{q}_c(\vx_i)= \frac{1}{Z_i}\sum_{\substack{ \vx_k \in \mathcal{N}_i}} \sI[\hat{y}_k = c] \cdot s_{ik} \qquad c \in [C]
\end{equation}
where $Z_i$ is the normalization factor such $\sum_{c} \hat{q}_c(\vx_i) = 1$.
Using these posterior probabilities, we select a maximum of $m$ most reliable image-label pairs for each class where $m$ is the $\delta$-quantile of $[a_1, a_2, \ldots, a_C]$. We do this to ensure a near-uniform distribution of samples across all classes. $a_1, a_2, \ldots, a_C$ are defined as follows.
\begin{equation} \label{eq:perclass_samples}
    a_c = \sum_{i=1}^{n} \sI \left[ 
                    c=\argmax_{c^\prime} \hat{q}_{c^\prime}(\vx_i)  \;\;\&\;\;c \in Y_i
                    \right],\;\;\;\; c \in [C]
\end{equation}
Thus, $a_c$ is the number of samples for which $c$ is the highest probable class according to the posterior probability and $c$ is also in the partial label of those samples. For instance, when $\delta=0$, $m = \min([a_1, a_2, \ldots, a_C])$.

A key insight is that, as training progresses, the per-class agreements $a_c$ also increase. This leads to an increase in $m$ and, consequently, the number of reliable pseudo-labels. The selection procedure of $m$ most reliable images per class is described below.

For class $c$, the images for which the posterior probability approximated via KNN $\hat{q}_c(\vx_i)$ is greater than the threshold $\gamma_c$ are chosen as reliable images. Thus, the set of reliable pairs belonging to the $c$-th class (denoted as $\mathcal{T}_c$) is defined as follows.
\begin{equation} \label{eq:reliable_set_c}
    \mathcal{T}_c =  \{ (\vx_i,c) \;|\; \hat{q}_c(\vx_i) > \gamma_c, c\in Y_i, i\in[n] \} \qquad c\in[C]
\end{equation}

For the $c$-th class, the threshold $\gamma_c$ is dynamically defined such that a maximum of $m$ samples can be selected per class. Note that, it is possible for a sample $\vx_i$ to satisfy the conditions for multiple different $\mathcal{T}_c$. In that case, we choose the one $\mathcal{T}_c$ that has the highest $\hat{q}_c(\vx_i)$. Finally, we form the reliable image-label pair set $\mathcal{T}$  as 
\begin{equation} \label{eq:reliable_set}
    \mathcal{T} = \bigcup^C_{c=1} \mathcal{T}_c = {\{(\vx_i,\Tilde{y}_i)\}}^{\Tilde{n}}_{i=1}
\end{equation}
where $\Tilde{y}_i$ is the reliable pseudo-label for $\vx_i$ and $\Tilde{n}$ is the number of selected reliable samples.

\begin{algorithm}[t]
    \caption{Pseudo-code of PALS}\label{alg:PALS}
    \begin{algorithmic}[1]
    \Require Dataset $\mathcal{D}$, encoder $f(\cdot)$, classifier $h(\cdot)$, epochs $T_{max}$, mini-batch size $B$, hyper-parameters $k, \delta, \zeta, r, \lambda_{max}, \lambda_{min}$
    \For{$t=1,2,\ldots,T_{max}$}
    \State Compute psuedo-labels $\hat{y}_i$ for all samples $\vx_i$ in the dataset using Eq.(\ref{eq:pseudolabel});
    \State Compute the posterior probability vectors using Eq.(\ref{eq:posterior_prob});
    \State Compute the maximum samples to be selected per class $m$ using Eq.(\ref{eq:perclass_samples});
    \State Build the Reliable image-label pairs set $\mathcal{T}$ using Eq.(\ref{eq:reliable_set_c});
    \State Shuffle $\mathcal{T}$ into $\frac{\mathcal{T}}{B}$ mini-batches;
    \For{$b = 1,2,\ldots,\frac{\mathcal{T}}{B}$}
    \State Update parameters of $f(\cdot)$ and $h(\cdot)$ by minimizing loss in Eq.(\ref{eq:final_loss});
    \EndFor
    \State Perform Partial label augmentation using Eq.(\ref{eq:partiallabel_augmentation});
    \EndFor
    \Ensure parameters of $f(\cdot)$ and $h(\cdot)$
    \end{algorithmic}
\end{algorithm}

\subsubsection{Noise robust learning} \label{sec:noise_robust_learning}
We train a noise-robust classifier $h(\cdot)$ by leveraging label smoothing~\cite{szegedy2016rethinking} and optional regularization components. Label smoothing uses a positively weighted average of the hard training labels and uniformly distributed soft labels to generate the smoothed labels. Let $\mathbf{e}_{\Tilde{y}_i}$ be $C$-dimensional one-hot encoding of label $\Tilde{y}_i$ such that its $\Tilde{y}_i^{th}$ element is one and rest of the entries are zero. The corresponding smoothed label $\mathbf{e}_{\Tilde{y}_i}^{LS,r}$ is:
\begin{equation}
    \mathbf{e}_{\Tilde{y}_i}^{LS,r} = (1-r)\cdot\mathbf{e}_{\Tilde{y}_i} + \frac{r}{C}\cdot\1
\end{equation}
Then the per-sample classification loss is formulated as:
\begin{equation}
    \begin{split}
        \mathcal{L}_{ce}(\vx_i,\mathbf{e}_{\Tilde{y}_i}^{LS,r}) = -(1-r)\log( h^{\Tilde{y}_i}(f(\vx_i)))\\
        -\frac{r}{C}\sum^C_{j=1} \log (h^j(f(\vx_i)))
    \end{split} 
\end{equation}
Although label smoothing introduces an additional hyper-parameter $r$, fixing it to a high value that is greater than the true noise rate ($r=0.5$ in the experiments) improves performance \cite{lukasik2020does}.

\subsubsection{Optional regularization components} In line with the prior works, we also enhance the performance of PALS by leveraging Mix-up \cite{zhang2017mixup} and Consistency regularization (CR) ~\cite{bachman2014learning}. Consider a pair of samples $(\vx_i,\mathbf{e}_{\Tilde{y}_i}^{LS,r})$ and $(\vx_j,\mathbf{e}_{\Tilde{y}_j}^{LS,r})$. We create the Mix-up image by linearly interpolating with a factor $\alpha_{mix} \sim Beta(\zeta, \zeta)$ where $\zeta$ is a parameter in the beta distribution and define the Mix-up loss below.
\begin{equation}
    \vx_{ij}^{mix} = \alpha_{mix}\vx_i + (1-\alpha_{mix})\vx_j
\end{equation}
\begin{equation}
    \begin{split}
        \mathcal{L}_{mix}(\vx_i,\vx_j) = \alpha_{m}\mathcal{L}_{ce}(\vx_{ij}^{mix},\mathbf{e}_{\Tilde{y}_i}^{LS,r})\\
        +  (1-\alpha_{m})\mathcal{L}_{ce}(\vx_{ij}^{mix},\mathbf{e}_{\Tilde{y}_j}^{LS,r})
    \end{split}
\end{equation}

Then, we include CR in the complete training objective. The idea is that the model should produce similar predictions for perturbed versions of the same image. We implement a variation by assigning the same pseudo-label to both weak and strong augmentations of an image.
\begin{equation} \label{eq:final_loss}
    \begin{split}
        \mathcal{L}_{final} = \sum_{i,j,j^\prime \in [\Tilde{n}]} \mathcal{L}_{mix}(aug_w(\vx_i),aug_w(\vx_j))\\
                        + \mathcal{L}_{mix}(aug_s(\vx_i),aug_s(\vx_{j\prime})) \qquad  i,j,j^\prime \in [\Tilde{n}]
    \end{split}
\end{equation}
where $aug_w(\cdot)$ and $aug_s(\cdot)$ represent weak and strong augmentation function respectively. Since Mix-up and CR are optional components, we show the effectiveness of our framework without these components in table~\ref{tab:no_cr_mix}.

\subsubsection{Partial label augmentation} \label{sec:partial_label_augmentation}
In Pseudo-labelling (Eq.\ref{eq:reliable_set_c}), for every sample image, we select its reliable pseudo-label only from the partial label. This limits the number of correct samples that can be selected in the NPLL case. To overcome this issue, we include the highest probability class of the current model prediction in the partial label for the next iteration if it is greater than a threshold $\lambda^t$. $\lambda^t$ is a decaying threshold and can be interpreted as balancing between precision and recall. Initially, the high threshold implies that few accurate predictions are used to avoid error propagation. The threshold is later decreased to allow more samples to be de-noised.
\begin{equation} \label{eq:partiallabel_augmentation}
    Y^{t+1}_i = \begin{cases}
        \begin{split}
            Y_i \cup \{ \argmax_c\; h_t^c(f_t(aug_w(\vx_i))) \},\\ 
        \text{ if } \max_c h_t^c(f_t(aug_w(\vx_i))) > \lambda^t
        \end{split}\\
        Y_i, \text{  otherwise}.
        
    \end{cases}
\end{equation}
Here, $f_t$ and $h_t$ are the encoder and classifier at the $t^{th}$ iteration. It is important to note that, we include the highest probability class in the true partial label set $Y_i$ to get $Y_i^{t+1}$. Unlike previous works \cite{lian2023irnet,xu2023alim}, we do not directly disambiguate the partial label. Instead, we merely include a possible candidate to allow its selection in the pseudo-labelling stage.
\section{Experiments}
\label{sec:experiments}

\subsection{Experimental Setup}
\textbf{Datasets} We construct the PLL and NPLL benchmark for four different datasets: CIFAR-10, CIFAR-100, CIFAR-100H~\cite{krizhevsky2009learning} and CUB-200~\cite{welinder2010caltech}. We follow the same dataset generation process of \cite{wang2022pico+}. The generation process is governed by two parameters: the partial rate ($q$) and the noise rate ($\eta$). The partial rate $q$ represents the probability of an incorrect label to be present in the candidate set and the noise rate $\eta$ represents the probability of excluding the correct label from the candidate set. For CIFAR10, we experiment with $q\in\{0.1,0.3,0.5\}$, and for other datasets (with a larger number of classes) we select $q$ from the set $\{0.01,0.03,0.05\}$. We conduct experiments using different values of $\eta\in\{0.1, 0.2, 0.3, 0.4, 0.5\}$, where values of $\eta$ exceeding 0.3 indicate high noise scenarios. Further, we show the effectiveness of our approach in real-world annotation settings by considering three fine-grained crowdsourced datasets: Treeversity\#6, Benthic, Plankton \cite{schmarje2022one}. \textit{Treeversity} is a publicly available dataset of plant images, crowdsourced by the Arnold Arboretum of Harvard University. \textit{Benthic} is a collection of seafloor images containing flora and fauna. \textit{Plankton} consists of underwater plankton images. Each image in these datasets has multiple annotations provided by domain experts. 


\textbf{Baselines} We compare PALS against nine prior SOTA methods, four PLL and five NPLL. The PLL methods include 1) RC~\cite{feng2020provably}: a risk-consistent classifier for PLL; 2) LWS~\cite{wen2021leveraged}: employs a loss trade-off between candidate labels and non-candidate labels; 3) PiCO~\cite{wang2022pico}: learns discriminative representations using contrastive and prototype learning 4) CRDPLL~\cite{wu2022revisiting}: adapts consistency regularization for PLL. The NPLL methods include 1) PiCO+~\cite{wang2022pico+}: extends PiCO for NPLL by including additional losses; 2) FREDIS~\cite{qiao2023fredis}: performs partial label disambiguation and refinement 3) IRNet~\cite{lian2023irnet}: detects and corrects the partial labels iteratively; 4) UPLLRS~\cite{Shi2023Unreliable}: proposes a two-stage framework for dataset separation and disambiguation; 3) ALIM~\cite{xu2023alim}: aggregates the partial labels and model outputs using two variants of normalization (Scale and OneHot).
If available, we directly reference the numbers from the existing literature, and whenever possible, we present results obtained by customizing the publicly available official code repository to suit the specific configuration. 

\textbf{Implementation Details.} To ensure a fair comparison, we use ResNet-18 \cite{he2016deep} as the encoder $f(\cdot)$ and $h(\cdot)$ as the linear classifier. Other methods either use ResNet-18 \cite{he2016deep} or a stronger backbone e.g. FREDIS uses ResNet32. Strong augmentation function utilizes autoaugment \cite{cubuk2018autoaugment} and cutout \cite{devries2017improved}. Faiss \cite{johnson2019billion} is used to speed up the nearest neighbour computation. Irrespective of the partial and noise rate, we set $K=15$, $\delta=0.25$, $\zeta=1.0$, $r=0.5$, and decrease $\lambda_t$ linearly in $[0.45,0.35]$.
For CIFAR datasets, we choose the SGD optimizer with a momentum of $0.9$ and set the weight decay to $0.001$. The initial learning rate is set to $0.1$ and decayed using a cosine scheduler, and the model is trained for $500$ epochs. For CUB-200, as per the existing works, we initialize the encoder with ImageNet-1K \cite{russakovsky2015imagenet} pre-trained weights. We choose the SGD optimizer with a momentum of $0.9$ and set the weight decay to $5e{-}4$. We set the initial learning rate to $0.05$ and decay it by $0.2$ using the step scheduler at epochs $[60,120,160,200]$ and train for $250$ epochs. On crowdsourced datasets, we utilize a ResNet50 backbone pre-trained on ImageNet-1K, to align with prior efforts. We report the mean and standard deviation of the accuracy on test sets based on three runs for each experiment.
All experiments are implemented with PyTorch and carried out on the NVIDIA GeForce RTX 2080 Ti GPU.

\begin{table*}[t]
\centering
\fontsize{8pt}{8pt}\selectfont
\setlength{\tabcolsep}{2pt}
\begin{tabular}{l|ccc|ccc|ccc}
\hline
\multirow{2}{*}{CIFAR-10~} &\multicolumn{3}{c|}{ $q=0.1$} &\multicolumn{3}{c|}{$q=0.3$} &\multicolumn{3}{c}{$q=0.5$}\\
\cline{2-10}
&$\eta=0.1$ &$\eta=0.2$ &$\eta=0.3$ &$\eta=0.1$ &$\eta=0.2$ &$\eta=0.3$ &$\eta=0.1$ &$\eta=0.2$ &$\eta=0.3$ \\
\hline
RC& 80.87$\pm$0.30& 78.22$\pm$0.23& 75.24$\pm$0.17& 79.69$\pm$0.37& 75.69$\pm$0.63& 71.01$\pm$0.54& 72.46$\pm$1.51& 59.72$\pm$0.42&49.74$\pm$0.70\\
 LWS& 82.97$\pm$0.24& 79.46$\pm$0.09& 74.28$\pm$0.79& 80.93$\pm$0.28& 76.07$\pm$0.38& 69.70$\pm$0.72& 70.41$\pm$2.68& 58.26$\pm$0.28&39.42$\pm$3.09\\ 
FREDIS& 90.57$\pm$0.23& 88.01$\pm$ 0.19 & 84.35$\pm$0.20& 89.02$\pm$0.15& 86.14$\pm$0.17 & 81.02$\pm$0.60& 87.42$\pm$0.21& 80.81$\pm$0.99 &65.15$\pm$0.13\\
PiCO& 90.78$\pm$0.24& 87.27$\pm$0.11& 84.96$\pm$0.12& 89.71$\pm$0.18& 85.78$\pm$0.23& 82.25$\pm$0.32& 88.11$\pm$0.29& 82.41$\pm$0.30&68.75$\pm$2.62 \\
CRDPLL& 93.48$\pm$0.17& 89.13$\pm$0.39& 86.19$\pm$0.48& 92.73$\pm$0.19& 86.96$\pm$0.21& 83.40$\pm$0.14& 91.10$\pm$0.07& 82.30$\pm$0.46&73.78$\pm$0.55 \\
 IRNet& 93.44$\pm$0.21& 92.57$\pm$0.25& 92.38$\pm$0.21& 92.81$\pm$0.19& 92.18$\pm$0.18& 91.35$\pm$0.08& 91.51$\pm$0.05& 90.76$\pm$0.10&86.19$\pm$0.41\\
PiCO+ & 94.48$\pm$0.02  & 94.74$\pm$0.13  & 94.43$\pm$0.19  & 94.02$\pm$0.03  & 94.03$\pm$0.01  & 92.94$\pm$0.24  & 93.56$\pm$0.08  & 92.65$\pm$0.26  & 88.21$\pm$0.37 \\
UPLLRS& 95.16$\pm$0.10& -& 94.65$\pm$0.23& 94.32$\pm$0.21& -& 93.85$\pm$0.31& 92.47$\pm$0.19& -&91.55$\pm$0.38\\
ALIM-Scale & 95.71$\pm$0.01  & 95.50$\pm$0.08  & 95.35$\pm$0.13  & 95.31$\pm$0.16  & 94.77$\pm$0.07  & 94.36$\pm$0.03  & 94.71$\pm$0.04  & 93.82$\pm$0.13  & 90.63$\pm$0.10 \\
ALIM-Onehot & 95.83$\pm$0.13  & 95.86$\pm$0.15  & 95.75$\pm$0.19  & 95.52$\pm$0.15  & 95.41$\pm$0.13  & 94.67$\pm$0.21  & 95.19$\pm$0.24  & 93.89$\pm$0.21  & 92.26$\pm$0.29  \\ 
 PALS & {\bf 96.28$\pm$0.05}  & {\bf 96.17$\pm$0.18}  & {\bf 95.90$\pm$0.20}  & {\bf 95.96$\pm$0.08}  & {\bf 95.76$\pm$0.12}  & {\bf 95.43$\pm$0.17}  & {\bf 95.52$\pm$0.13}  & {\bf 95.89$\pm$0.14}  & {\bf 94.18$\pm$0.10}  \\ 
\hline
\end{tabular}
\begin{tabular}{l|ccc|ccc|ccc}
\hline
\multirow{2}{*}{CIFAR-100} &\multicolumn{3}{c|}{ $q=0.01$} &\multicolumn{3}{c|}{$q=0.03$} &\multicolumn{3}{c}{$q=0.05$}\\
\cline{2-10}
&$\eta=0.1$ &$\eta=0.2$ &$\eta=0.3$ &$\eta=0.1$ &$\eta=0.2$ &$\eta=0.3$ &$\eta=0.1$ &$\eta=0.2$ &$\eta=0.3$ \\
\hline
RC& 52.73$\pm$1.05& 48.59$\pm$1.04& 45.77$\pm$0.31& 52.15$\pm$0.19& 48.25$\pm$0.38& 43.92$\pm$0.37& 46.62$\pm$0.34& 45.46$\pm$0.21&40.31$\pm$0.55\\
 LWS& 56.05$\pm$0.20& 50.66$\pm$0.59& 45.71$\pm$0.45& 53.59$\pm$0.45& 48.28$\pm$0.44& 42.20$\pm$0.49& 45.46$\pm$0.44& 39.63$\pm$0.80&33.60$\pm$0.64\\
  FREDIS& 64.73$\pm$0.28& 64.53 $\pm$ 0.39 & 59.42$\pm$0.18& 67.38$\pm$0.40 & 63.86$\pm$0.50 & 60.86$\pm$0.72 & 66.43$\pm$0.22& 63.09 $\pm$ 0.22 &56.15$\pm$0.18\\
 PiCO& 68.27$\pm$0.08& 62.24$\pm$0.31& 58.97$\pm$0.09& 67.38$\pm$0.09& 62.01$\pm$0.33& 58.64$\pm$0.28& 67.52$\pm$0.43& 61.52$\pm$0.28&58.18$\pm$0.65\\
 CRDPLL& 68.12$\pm$0.13& 65.32$\pm$0.34& 62.94$\pm$0.28& 67.53$\pm$0.07& 64.29$\pm$0.27& 61.79$\pm$0.11& 67.17$\pm$0.04& 64.11$\pm$0.42&61.03$\pm$0.43\\
 IRNet& 71.17$\pm$0.14& 70.10$\pm$0.28& 68.77$\pm$0.28& 71.01$\pm$0.43& 70.15$\pm$0.17& 68.18$\pm$0.30& 70.73$\pm$0.09& 69.33$\pm$0.51&68.09$\pm$0.12\\
PiCO+ & 75.04$\pm$0.18  & 74.31$\pm$0.02  & 71.79$\pm$0.17  & 74.68$\pm$0.19  & 73.65$\pm$0.23  & 69.97$\pm$0.01  & 73.06$\pm$0.16  & 71.37$\pm$0.16  & 67.56$\pm$0.17 \\
 UPLLRS& 75.73$\pm$0.41& -& 71.72$\pm$0.39& -& -& -& 74.73$\pm$0.24& -&70.31$\pm$0.22\\
ALIM-Scale & 77.37$\pm$0.32  & 76.81$\pm$0.05  & 76.45$\pm$0.30  & 77.60$\pm$0.18  & 76.63$\pm$0.19  & 75.92$\pm$0.14  & 76.86$\pm$0.23  & 76.44$\pm$0.12  & 75.67$\pm$0.17 \\
ALIM-Onehot & 76.52$\pm$0.19  & 76.55$\pm$0.24  & 76.09$\pm$0.23  & 77.27$\pm$0.23  & 76.29$\pm$0.41  & 75.29$\pm$0.57  & 76.87$\pm$0.20  & 75.23$\pm$0.42  & 74.49$\pm$0.61  \\ 
PALS & {\bf 80.90$\pm$0.50}  & {\bf 80.45$\pm$0.49}  & {\bf 79.78$\pm$0.13}  & {\bf 80.33$\pm$0.04}  & {\bf 79.40$\pm$0.50}  & {\bf 78.52$\pm$0.24}  & {\bf 80.00$\pm$0.25}  & {\bf 79.08$\pm$0.22}  & {\bf 77.87$\pm$0.29} \\
\hline
\end{tabular}
\caption{Comparison of PALS with the previous state-of-the-art methods across various partial and noise rates. `-' indicates missing values due to the code being unavailable.}
\label{tab:cifar}
\end{table*}

\begin{table}[]
    \centering
    \fontsize{8pt}{8pt}\selectfont
    \begin{tabular}{l|c|c|c}
    \hline
       CIFAR-100  & $q=0.01$ & $q=0.05$ & $q=0.1$ \\
    \hline
        RC & 75.36$\pm$0.06& 74.44$\pm$0.31& 73.79$\pm$0.29\\
        LWS & 64.55$\pm$1.98& 50.19$\pm$0.34& 44.93$\pm$1.09\\
        PiCO & 73.78$\pm$0.15& 72.78$\pm$0.38& 71.55$\pm$0.31\\
        CRDPLL & 79.74$\pm$0.07& 78.97$\pm$0.13& 78.51$\pm$0.24\\
        PiCO+ &76.29$\pm$0.42 & 76.17$\pm$0.18 & 75.55$\pm$0.21 \\
        PALS &{\bf 81.46$\pm$0.13} & {\bf 81.00$\pm$0.26} & {\bf 80.77$\pm$0.12} \\
    \hline  
    \end{tabular}  
    \caption{PLL Experiments. $\eta$ is set to $0$.}
    \label{tab:no_noise}
\end{table}

\begin{table}[]
    \centering  
    \fontsize{8pt}{8pt}\selectfont
    \begin{tabular}{l|c|c}
    \hline
        \multirow{2}{*}{CIFAR-100} &$q=0.05$  &$q=0.05$ \\
         &$\eta=0.4$  &$\eta=0.5$ \\
        \hline
        FREDIS & 53.44$\pm$0.39 & 48.05$\pm$0.20  \\
         PiCO& 44.17$\pm$0.08&35.51$\pm$1.14\\
         CRDPLL& 57.10$\pm$0.24&52.10$\pm$0.36\\
        UPLLRS & -& 64.78$\pm$0.53  \\
        PiCO+ & 66.41$\pm$0.58   & 60.50$\pm$0.99  \\
        ALIM-Scale & 74.98$\pm$0.16  & 72.26$\pm$0.25  \\
        ALIM-Onehot & 71.76$\pm$0.56  & 69.59$\pm$0.62  \\
        PALS & {\bf 76.72$\pm$0.41}  & {\bf 74.79$\pm$0.40}  \\
    \hline
    \end{tabular}
    \caption{Extreme noise experiments. $\eta$ values are above $0.3$}
    \label{tab:extreme_noise}
\end{table}

\begin{table*}[t]
    \centering  
    \fontsize{8pt}{8pt}\selectfont
    \begin{tabular}{l|c|c|c|c|c}
    \hline
        Dataset &CIFAR-100H &CUB-200 &CUB-200 &CUB-200 &CUB-200\\ 
        $q$ &$0.5$ &$0.05$ &$0.01$ &$0.05$ &$0.1$\\
        $\eta$ &$0.2$ &$0.2$ &$0.0$ &$0.0$ &$0.0$ \\
        \hline
        PiCO & 59.8$\pm$0.25& 53.05$\pm$2.03& 74.14$\pm$0.24  & 72.17$\pm$0.72  & 62.02$\pm$1.16  \\
        PiCO+ & 68.31$\pm$0.47  &  60.65$\pm$0.79  & 69.83$\pm$0.07& 72.05$\pm$0.80& 58.68$\pm$0.30\\
        ALIM-Scale & 73.42$\pm$0.18  & 68.38$\pm$0.47  & 74.44$\pm$0.68 & 74.26$\pm$0.32 & 62.88$\pm$0.90 \\
        ALIM-Onehot & 72.36$\pm$0.20  & 63.91$\pm$0.35  & -& -& -\\
        PALS & {\bf 75.16$\pm$0.59}  &  {\bf 75.76$\pm$0.27}  & {\bf 80.71$\pm$0.21}  & {\bf 79.46$\pm$0.25}  & {\bf 78.73$\pm$0.36}  \\
        \hline
    \end{tabular}
    \caption{Simulated fine-grained datasets.}
    \label{tab:fine-grained}
\end{table*}

\begin{table}[t]
    \centering
    \fontsize{8pt}{8pt}\selectfont
    \begin{subtable}{1.\linewidth}
    \begin{tabular}{l|c|c|c}
    \hline
         & Treeversity & Benthic & Plankton \\
    \hline
    Divide-Mix & 77.84$\pm$0.52 & 71.72$\pm$1.21 & 91.79$\pm$0.51 \\
    ELR+ & 79.05$\pm$1.22 &  67.53$\pm$1.76 & 91.76$\pm$0.91 \\
    PiCO & 84.74$\pm$0.42 & 51.01$\pm$0.74 & 24.95$\pm$0.13 \\
    PALS & 81.56$\pm$0.28 & 77.50$\pm$0.35 & 90.04$\pm$0.50 \\
    \hline
    \end{tabular}
    \caption{Ten human annotators}
    \label{tab:crowd_sourced_10annot}
    \end{subtable}
    \newline
    \vspace*{0.125 em}
    \newline
    \begin{subtable}{1.\linewidth}
    \begin{tabular}{l|c|c|c}
    \hline
         & Treeversity & Benthic & Plankton \\
    \hline
    Divide-Mix & 76.38$\pm$1.78 & 71.50$\pm$0.42 & 91.96$\pm$0.52 \\
    ELR+ & 77.07$\pm$0.43 &  69.35$\pm$0.97 & 91.36$\pm$0.40 \\
    ALIM-Onehot & 82.58$\pm$0.92 & 49.17$\pm$3.06 & 23.75$\pm$2.53 \\
    PALS & 80.47$\pm$1.22 & 76.52$\pm$0.03 & 90.20$\pm$0.97 \\
    \hline
    \end{tabular}
    \caption{Three human annotators}
    \label{tab:crowd_sourced_3annot}
    \end{subtable}
    \caption{Fine-grained crowdsourced Datasets}
    \label{tab:crowd_sourced}
\end{table}

\subsection{Results and Discussion}

\textbf{SOTA comparisons:} Table~\ref{tab:cifar} compares PALS with the SOTA methods on CIFAR-10 and CIFAR-100 datasets. The results are presented with three different partial rates and three different noise rates, comprising a total of nine combinations. PALS surpasses all other methods in all settings, providing clear evidence of its effectiveness. It is also apparent that the PLL approaches struggle to generalize effectively in noisy settings, highlighting the necessity for noise-resistant adaptations.

The performance of all methods drops with an increase in the size of the candidate label set (higher $q$) and with an increase in noise rate. A key aspect is to observe the performance across the first and last column of Table~\ref{tab:cifar} (the easiest vs. the most complex setting). In several methods, the performance significantly drops; for instance, on CIFAR-10, the performance of LWS drops from 82.97 to 39.42. PALS and ALIM fare better in this regard and are able to retain their performance with increased ambiguity. Another noteworthy observation is that the performance improvement of PALS becomes more pronounced in CIFAR-100 compared to the CIFAR-10 dataset. 

\textbf{PLL results:} While our primary focus is on NPLL, assessing performance in the noise-free setting (PLL) is crucial to ensure that the adaptations for noise do not negatively affect the zero noise performance. 
As most NPLL methods do not provide results in noise-free settings, we limit our comparisons primarily to the PLL methods. In Table~\ref{tab:no_noise}, we evaluate PALS on the CIFAR-100 PLL benchmark. The results indicate that PALS also achieves superior performance in the absence of noise.

\textbf{Extreme noise results:} In Table \ref{tab:extreme_noise}, we present the results in extremely noisy scenarios (e.g., $\eta=0.5$, the correct label is absent from half of the partial labels). For most methods, the performance drastically drops. For example, for PiCO+ the performance drops from $76.17$ ($\eta=0$) to $60.50$ ($\eta=0.5$). For CRDPLL, the drop is from $78.97$ ($\eta=0$) to $52.10$ ($\eta=0.5$). ALIM and PALS fare well here, with PALS obtaining the best results.

\textbf{Simulated fine-grained datasets:} We first evaluate PALS on the CIFAR-100H dataset, which considers label hierarchy and divides the CIFAR-100 dataset into 20 coarse labels (5 fine-grained labels per coarse label). When curating the NPLL benchmark, the candidate sets are restricted to lie within the same coarse label. PALS obtains the best results in this modified setting (Table~\ref{tab:fine-grained}). 

We also evaluate on the CUB-200 bird classification dataset. Since all the images exclusively showcase birds, the candidate labels are intricately linked, presenting a heightened challenge for disambiguation. Table~\ref{tab:fine-grained} presents the results with different $q$ and $\eta$ values. PALS outperforms other approaches by a significant margin, bringing over 7pp improvement over ALIM-Scale at $q=0.05$ and $\eta=0.2$.

\textbf{Fine-grained crowd-sourced datasets:} Prior efforts of classification on crowd-sourced datasets~\cite{schmarje2022one} take a data-centric approach, involving label enhancement through semi-supervised and noisy label learning, with subsequent evaluation by training a model on the enhanced dataset. Instead, we tackle this as a PLL/NPLL problem and employ PALS. As the only key change in the framework, we utilize the normalized annotation label frequency per image in the weighted KNN step. We construct the partial label by taking into account the annotations provided by human annotators. Table~\ref{tab:crowd_sourced_10annot} presents results under the assumption of ten available human annotators, simulating the PLL scenario with a negligible noise rate, while Table~\ref{tab:crowd_sourced_3annot} displays results assuming three human annotators, simulating the NPLL scenario characterized by a substantial noise rate. We compare against two noisy label learning methods: Divide-Mix~\cite{li2020dividemix} and ELR+~\cite{liu2020early}, a PLL method (PiCO) and an NPLL method (ALIM). In all cases, we use a pre-trained ResNet50 backbone to allow a fair one-to-one comparison. PALS achieves competitive performance on Treeversity.
On Benthic and Plankton, which contain underwater images, there is a sharp decline in the performance of PiCO and ALIM. PALS displays promising generalization capability by outperforming them by a significant margin.


\begin{table}[]
    \centering
    \setlength{\tabcolsep}{6pt}
    \scriptsize
    \begin{tabular}{l|cc|cc}
    \hline
    \multirow{2}{*}{CIFAR-10} &\multicolumn{2}{c|}{ $q=0.1$} &\multicolumn{2}{c}{$q=0.5$} \\
    \cline{2-5}
          & $\eta=0.1$  &$\eta=0.3 $ &$\eta=0.1 $  &$\eta=0.3$ \\
    \hline
        PiCO & 90.78$\pm$0.24 & 84.96$\pm$0.12 &88.11$\pm$0.29& 68.75$\pm$2.62\\
        PiCO+ & 93.64$\pm$0.19 & 92.18$\pm$0.38&91.07$\pm$0.02& 84.08$\pm$0.42\\
        ALIM-Scale & 94.15$\pm$0.14 & 93.28$\pm$0.08& 92.52$\pm$0.12& 86.51$\pm$0.21\\
        ALIM-Onehot & 94.15$\pm$0.15 & 93.77$\pm$0.27& 92.67$\pm$0.12& 89.80$\pm$0.38\\
        PALS & 94.77$\pm$0.12 & 94.53$\pm$0.08 & 93.71$\pm$0.06 & 92.08$\pm$0.55 \\
    \hline
    \end{tabular}
    \begin{tabular}{l|cc|cc}
    \hline
    \multirow{2}{*}{CIFAR-100} &\multicolumn{2}{c|}{ $q=0.03$} &\multicolumn{2}{c}{$q=0.05$} \\
    \cline{2-5}\textbf{}
          & $\eta=0.1$  &$\eta=0.3 $ &$\eta=0.3 $  &$\eta=0.5 $ \\
    \hline
        PiCO & 67.38$\pm$0.09& 58.64$\pm$0.28&58.18$\pm$0.60& 35.51$\pm$1.14\\
        PiCO+ & 70.89$\pm$0.13& 64.22$\pm$0.23&62.24$\pm$0.38& 45.29$\pm$0.14\\
        ALIM-Scale & 74.33$\pm$0.11& 71.69$\pm$0.39& 71.68$\pm$0.14& 64.74$\pm$0.16\\
        ALIM-Onehot & 71.43$\pm$0.21& 70.14$\pm$0.25& 70.05$\pm$0.43& 63.39$\pm$0.82\\
        PALS & 76.68$\pm$0.08 & 75.00$\pm$0.16 & 73.56$\pm$0.26& 70.85$\pm$0.49\\
    \hline
    \end{tabular}
    \caption{Comparison of PALS with the previous state-of-the-art methods when excluding Mix-up and Consistency regularization from all the methods.}
    \label{tab:no_cr_mix}
\end{table}

\begin{table}[t]
\scriptsize
\setlength{\tabcolsep}{3pt}
\begin{subtable}{.5\linewidth}
\centering
    \begin{tabular}{cccc}
    \hline
        \multirow{2}{*}{Mix-up} & \multirow{2}{*}{CR} &$q=0.05$ &$q=0.05$ \\
        & &$\eta=0.3$ &$\eta=0.5$ \\

        \hline
        \xmark & \xmark & 73.30 & 71.41 \\
        \xmark & \cmark & 74.33 & 72.51 \\
        \cmark & \xmark & 75.91 & 73.12 \\
        \cmark & \cmark & 77.61 & 75.15 \\
    \hline
    \end{tabular}
\caption{Influence of regularization}
\label{tab:ablation_mix_cr}
\end{subtable}%
\begin{subtable}{.5\linewidth}
    \centering    
    \begin{tabular}{lcc}
    \hline
        \multirow{2}{*}{K}  & $q=0.05$ &$q=0.05$ \\
        &$\eta=0.3$  &$\eta=0.5 $ \\
        \hline
        10 & 77.87 & 74.51 \\
        15 & 77.61 & 75.15 \\
        25 & 78.29 & 74.98 \\
        50 & 77.01 & 75.60 \\
    \hline
    \end{tabular}
    \label{tab:ablation_K}
    \caption{Impact of K}
\end{subtable}
\newline
\vspace*{0.125 em}
\newline
\begin{subtable}{.5\linewidth}
    \centering     
    \begin{tabular}{lcccc}
    \hline
      $r$  &$0.6$ &$0.5$ &$0.4$ &$0.0$\\
        \hline
        $q{=}0.05$&\multirow{2}{*}{76.87} &\multirow{2}{*}{77.61} & \multirow{2}{*}{78.19} & \multirow{2}{*}{75.64} \\
        $\eta{=}0.3$ & & & & \\
        \hline
        $q{=}0.05$ & \multirow{2}{*}{74.17} & \multirow{2}{*}{75.15}  &\multirow{2}{*}{74.41} & \multirow{2}{*}{65.63} \\
        $\eta{=}0.5$ & & & & \\
    \hline
    \end{tabular}
    \caption{Impact of label smoothing rate $r$}
    \label{tab:ablation_ls}
\end{subtable}%
\begin{subtable}{.5\linewidth}
    \centering    
    \begin{tabular}{lcccc}
    \hline
        $\delta$&$0.0$ &$0.25$ &$0.50$ &$0.75$\\
        \hline
        $q{=}0.05$ & \multirow{2}{*}{75.92} & \multirow{2}{*}{77.61} & \multirow{2}{*}{78.22} & \multirow{2}{*}{77.62}\\
        $\eta{=}0.3$ & & & & \\
        \hline
        $q{=}0.05$ & \multirow{2}{*}{73.50} & \multirow{2}{*}{75.15} & \multirow{2}{*}{74.72} & \multirow{2}{*}{72.24}\\
        $\eta{=}0.5$ & & & & \\
    \hline
    \end{tabular}
    \caption{Impact of $\delta$}
    \label{tab:ablation_delta}
\end{subtable}
\newline
\vspace*{0.125 em}
\newline
\begin{subtable}{.6\linewidth}
    \centering
    \begin{tabular}{lcc}
    \hline
        \multirow{2}{*}{Backbone} & $q=0.05$ &$q=0.05$ \\
        &$\eta=0.3$  &$\eta=0.5$ \\
        \hline
        Wide-ResNet-34x10 & 81.3 & 78.18 \\
        ResNext-50 & 79.55 & 76.37 \\
        SE-ResNet-50 & 79.46 & 75.56 \\
        DenseNet-121 & 78.75 & 74.36 \\
        ResNet-18 & 77.61 & 75.15 \\
    \hline
    \end{tabular}
    \caption{Influence of backbones}
    \label{tab:ablation_arch}
\end{subtable}%
\begin{subtable}{.4\linewidth}
    \centering
    \begin{tabular}{lcc}
    \hline
        Pseudo & $q=0.05$ &$q=0.05$ \\
        label &$\eta=0.3$  &$\eta=0.5$ \\
        \hline
        \xmark & 76.71 & 72.86 \\
        \cmark & 77.61 & 75.15 \\
    \hline
    \end{tabular}
    \caption{Impact of \cref{eq:pseudolabel}}
    \label{tab:ablation_pseudolabel}
\end{subtable}
\caption{Ablation studies}
\label{tab:ablation_studies}
\end{table}

\subsection{Ablation studies}
In Table~\ref{tab:ablation_studies}, we perform ablations using two different noise rates: ($\eta=0.3$ and $\eta=0.5$), while setting $q=0.05$. All ablations are performed using the CIFAR-100 dataset. 

\textbf{Influence of regularization components:} In table~\ref{tab:ablation_mix_cr}, we study the impact of using Mix-up and CR. Removing Mix-up causes a decrease of around $2-3$pp, and the exclusion of CR causes a decrease of about $1$pp, which validates their role in our framework. We observe that their combined usage generates a more substantial effect. It must be noted that prior methods also utilize Mix-up, CR or both. For a fair comparison, we remove both Mix-up and CR from all the approaches and present the results in table~\ref{tab:no_cr_mix}. PALS achieves SOTA results in this setup as well and in the high noise setting of CIFAR-100 ($\eta=0.5$), outperforms the second best method (ALIM-Scale) by a significant margin of over 6pp.

\begin{figure}[t!]
    \centering
    \includegraphics[width=1.\columnwidth]{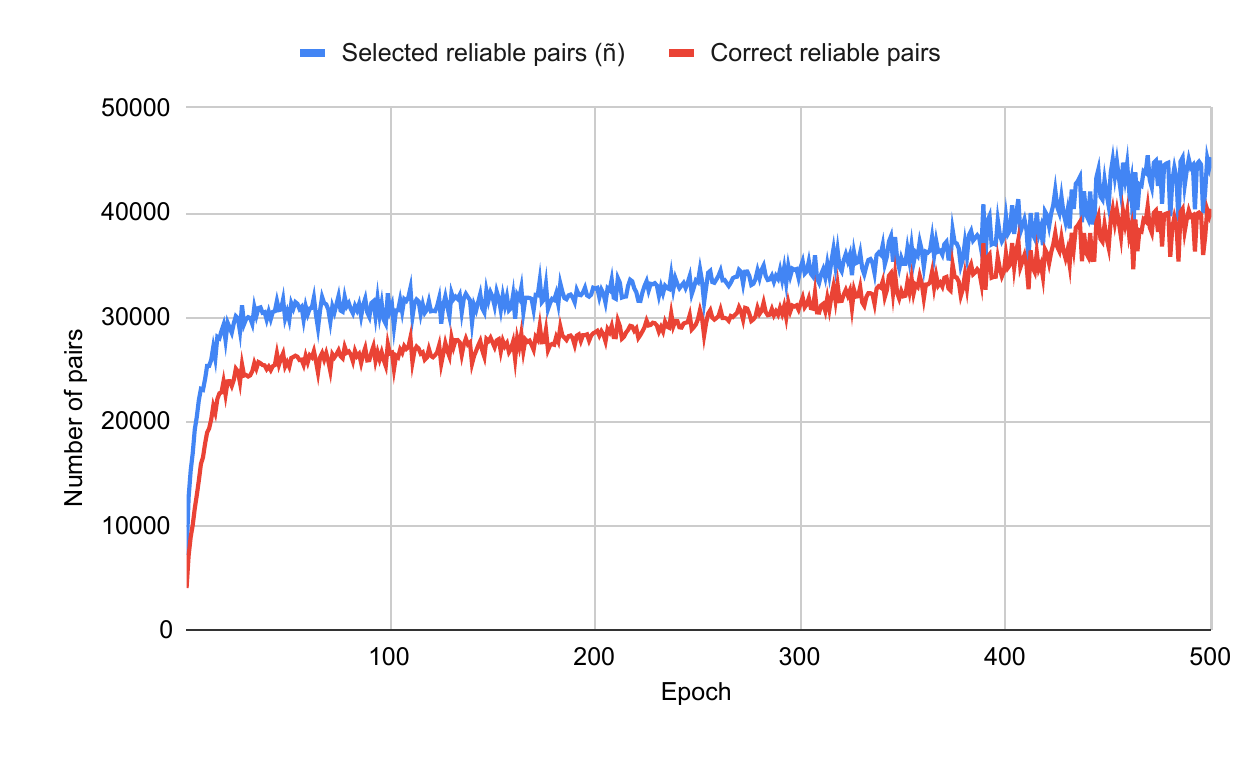}
    \caption{Number of selected reliable pairs ($\Tilde{n}$) while training on CIFAR-100 with $q=0.05$ and $\eta=0.3$}
    \label{fig:reliable_pairs}
\end{figure}

\textbf{Impact of $K$ in KNN:} We vary the value of $K$ from the set $\{10,15,25,50\}$. We find that performance remains similar across different $K$ values, with statistically insignificant differences. We use $K=15$ for all our experiments.

\textbf{Impact of $\delta$ parameter:} $\delta$ controls the number of samples chosen per class to construct the reliable set. In table~\ref{tab:ablation_delta}, $\delta=0.5$ yields the best results for $\eta=0.3$, while $\delta=0.25$ produces the best results for $\eta=0.5$. Lower $\delta$ implies that a smaller but more confident reliable set is constructed, which leads to better performance in a high noise setting.

\textbf{Impact of label smoothing:} In Table~\ref{tab:ablation_ls}, $r=0.4$ yields the best results for $\eta=0.3$, while $r=0.5$ produces the best results for $\eta=0.5$. Considering that elevated noise levels lead to increased noise in pseudo-labelling, a higher label smoothing rate appears to confer advantages. Moreover, when $r=0$, there is a significant drop in the accuracy under the extreme noise setting. The result strongly establishes the noise-robustness imparted by label smoothing in the NPLL context.

\textbf{Influence of backbones:} Table~\ref{tab:ablation_arch} establishes that PALS is robust to the choice of backbone. Prior effort~\cite{wu2022revisiting} has shown that this is not necessarily the case with other methods like PiCO.

\textbf{Impact of pseudo-label computation in \cref{eq:pseudolabel}:} The first row of \cref{tab:ablation_pseudolabel} shows the results when \cref{eq:pseudolabel} is skipped, and class posterior probabilities are computed directly based on the partial label $Y_k$. There is a small drop in performance, showing that initial pseudo-label computation is important for robustness against noise.

\textbf{Insights into pseudo-labelling:} Figure~\ref{fig:reliable_pairs} shows the number of selected reliable samples $\Tilde{n}$ (blue curve), while training on CIFAR-100 with $q=0.05$ and $\eta=0.3$. For visualization purposes, we also plot the number of the selected examples that are correct (red curve) by comparing them with the ground truth. In the first epoch, we observe that out of the 50000 samples, 7211 samples are selected for training (of which 4036 are correct). This highlights the effectiveness of the pseudo-labelling approach, given that the backbone is randomly initialized and there is a presence of significant noise in the partial labels. Since a good portion of selected samples are correct, this aids in learning improved features. Over time, both the number of selected examples and the proportion of correct labels increase, enhancing the quality of the feature representations. At convergence, ($\Tilde{n}$) is 45285, of which 40330 pseudo-labels are correct.

\textbf{Training time:} PALS is about 1.5 times faster to train compared to PiCO+ and ALIM. On an NVIDIA GeForce RTX 2080 Ti GPU, one epoch of PALS takes 40 seconds, while one epoch of ALIM takes 60 seconds. The numbers reported in our paper are by training for 500 epochs, while ALIM reports numbers after training for 800-1000 epochs (based on their official code). So effectively, PALS is about three times faster to train. On larger models, the gap widens. For example, for a WideResNet, PALS takes 221 seconds, and ALIM takes 550 seconds per epoch, making PALS 2 times faster.

\section{Conclusion}

We propose PALS, a framework for NPLL that iteratively performs pseudo-labelling and classifier training. Pseudo-labelling involves the weighted KNN algorithm, and a classifier is trained using the labelled samples while leveraging label smoothing and optional regularization components. PALS achieves SOTA results in a comprehensive set of experiments, varying partial rates and noise rates. The gains are more pronounced with an increase in noise and the number of classes. Notably, PALS excels in the simulated fine-grained classification tasks. Unlike many previous methods, PALS preserves its performance as the ambiguity in the dataset increases. PALS also generalizes well to the real world, fine-grained, crowd-sourced datasets.

We provide thorough ablation studies to examine the functioning of the PALS framework. Our analysis underscores the consistent contributions of both consistency regularization and Mix-up to overall performance improvements. Additionally, we highlight the enormous benefits of label smoothing, particularly in high-noise scenarios. We believe that the PALS framework and the insights presented in this study would be helpful for the research community across a wide range of practical applications.

{
    \small
    \bibliographystyle{ieeenat_fullname}
    \bibliography{main}

\begin{thebibliography}{36}
\providecommand{\natexlab}[1]{#1}
\providecommand{\url}[1]{\texttt{#1}}
\expandafter\ifx\csname urlstyle\endcsname\relax
  \providecommand{\doi}[1]{doi: #1}\else
  \providecommand{\doi}{doi: \begingroup \urlstyle{rm}\Url}\fi

\bibitem[Bachman et~al.(2014)Bachman, Alsharif, and Precup]{bachman2014learning}
Philip Bachman, Ouais Alsharif, and Doina Precup.
\newblock Learning with pseudo-ensembles.
\newblock \emph{Advances in neural information processing systems}, 27, 2014.

\bibitem[Briggs et~al.(2012)Briggs, Fern, and Raich]{briggs2012rank}
Forrest Briggs, Xiaoli~Z Fern, and Raviv Raich.
\newblock Rank-loss support instance machines for miml instance annotation.
\newblock In \emph{Proceedings of the 18th ACM SIGKDD international conference on Knowledge discovery and data mining}, pages 534--542, 2012.

\bibitem[Cour et~al.(2009)Cour, Sapp, Jordan, and Taskar]{cour2009learning}
Timothee Cour, Benjamin Sapp, Chris Jordan, and Ben Taskar.
\newblock Learning from ambiguously labeled images.
\newblock In \emph{2009 IEEE Conference on Computer Vision and Pattern Recognition}, pages 919--926. IEEE, 2009.

\bibitem[Cubuk et~al.(2019)Cubuk, Zoph, Mane, Vasudevan, and Le]{cubuk2018autoaugment}
Ekin~D Cubuk, Barret Zoph, Dandelion Mane, Vijay Vasudevan, and Quoc~V Le.
\newblock Autoaugment: Learning augmentation strategies from data.
\newblock In \emph{Proceedings of the IEEE/CVF conference on computer vision and pattern recognition}, pages 113--123, 2019.

\bibitem[DeVries and Taylor(2017)]{devries2017improved}
Terrance DeVries and Graham~W Taylor.
\newblock Improved regularization of convolutional neural networks with cutout.
\newblock \emph{arXiv preprint arXiv:1708.04552}, 2017.

\bibitem[Feng et~al.(2020)Feng, Lv, Han, Xu, Niu, Geng, An, and Sugiyama]{feng2020provably}
Lei Feng, Jiaqi Lv, Bo Han, Miao Xu, Gang Niu, Xin Geng, Bo An, and Masashi Sugiyama.
\newblock Provably consistent partial-label learning.
\newblock \emph{Advances in neural information processing systems}, 33:\penalty0 10948--10960, 2020.

\bibitem[He et~al.(2016)He, Zhang, Ren, and Sun]{he2016deep}
Kaiming He, Xiangyu Zhang, Shaoqing Ren, and Jian Sun.
\newblock Deep residual learning for image recognition.
\newblock In \emph{Proceedings of the IEEE conference on computer vision and pattern recognition}, pages 770--778, 2016.

\bibitem[H{\"u}llermeier and Beringer(2005)]{hullermeier2005learning}
Eyke H{\"u}llermeier and J{\"u}rgen Beringer.
\newblock Learning from ambiguously labeled examples.
\newblock In \emph{International Symposium on Intelligent Data Analysis}, pages 168--179. Springer, 2005.

\bibitem[Jin and Ghahramani(2002)]{Jin2002learning}
Rong Jin and Zoubin Ghahramani.
\newblock Learning with multiple labels.
\newblock In \emph{Advances in Neural Information Processing Systems}. MIT Press, 2002.

\bibitem[Johnson et~al.(2019)Johnson, Douze, and J{\'e}gou]{johnson2019billion}
Jeff Johnson, Matthijs Douze, and Herv{\'e} J{\'e}gou.
\newblock Billion-scale similarity search with {GPUs}.
\newblock \emph{IEEE Transactions on Big Data}, 7\penalty0 (3):\penalty0 535--547, 2019.

\bibitem[Krizhevsky et~al.(2009)Krizhevsky, Hinton, et~al.]{krizhevsky2009learning}
Alex Krizhevsky, Geoffrey Hinton, et~al.
\newblock Learning multiple layers of features from tiny images.
\newblock 2009.

\bibitem[Li et~al.(2019)Li, Socher, and Hoi]{li2020dividemix}
Junnan Li, Richard Socher, and Steven~CH Hoi.
\newblock Dividemix: Learning with noisy labels as semi-supervised learning.
\newblock In \emph{International Conference on Learning Representations}, 2019.

\bibitem[Lian et~al.(2023)Lian, Xu, Chen, Sun, Liu, and Tao]{lian2023irnet}
Zheng Lian, Mingyu Xu, Lan Chen, Licai Sun, Bin Liu, and Jianhua Tao.
\newblock Irnet: Iterative refinement network for noisy partial label learning, 2023.

\bibitem[Liu et~al.(2020)Liu, Niles-Weed, Razavian, and Fernandez-Granda]{liu2020early}
Sheng Liu, Jonathan Niles-Weed, Narges Razavian, and Carlos Fernandez-Granda.
\newblock Early-learning regularization prevents memorization of noisy labels.
\newblock \emph{Advances in neural information processing systems}, 33:\penalty0 20331--20342, 2020.

\bibitem[Lukasik et~al.(2020)Lukasik, Bhojanapalli, Menon, and Kumar]{lukasik2020does}
Michal Lukasik, Srinadh Bhojanapalli, Aditya Menon, and Sanjiv Kumar.
\newblock Does label smoothing mitigate label noise?
\newblock In \emph{International Conference on Machine Learning}, pages 6448--6458. PMLR, 2020.

\bibitem[Lv et~al.(2020)Lv, Xu, Feng, Niu, Geng, and Sugiyama]{lv2020progressive}
Jiaqi Lv, Miao Xu, Lei Feng, Gang Niu, Xin Geng, and Masashi Sugiyama.
\newblock Progressive identification of true labels for partial-label learning.
\newblock In \emph{International conference on machine learning}, pages 6500--6510. PMLR, 2020.

\bibitem[Lv et~al.(2023)Lv, Liu, Feng, Xu, Xu, An, Niu, Geng, and Sugiyama]{lv2023robustness}
Jiaqi Lv, Biao Liu, Lei Feng, Ning Xu, Miao Xu, Bo An, Gang Niu, Xin Geng, and Masashi Sugiyama.
\newblock On the robustness of average losses for partial-label learning.
\newblock \emph{IEEE Transactions on Pattern Analysis and Machine Intelligence}, 2023.

\bibitem[Ortego et~al.(2021)Ortego, Arazo, Albert, O'Connor, and McGuinness]{ortego2021multi}
Diego Ortego, Eric Arazo, Paul Albert, Noel~E O'Connor, and Kevin McGuinness.
\newblock Multi-objective interpolation training for robustness to label noise.
\newblock In \emph{Proceedings of the IEEE/CVF Conference on Computer Vision and Pattern Recognition}, pages 6606--6615, 2021.

\bibitem[Qiao et~al.(2023)Qiao, Xu, Lv, Ren, and Geng]{qiao2023fredis}
Congyu Qiao, Ning Xu, Jiaqi Lv, Yi Ren, and Xin Geng.
\newblock Fredis: A fusion framework of refinement and disambiguation for unreliable partial label learning.
\newblock In \emph{International Conference on Machine Learning}, pages 28321--28336. PMLR, 2023.

\bibitem[Russakovsky et~al.(2015)Russakovsky, Deng, Su, Krause, Satheesh, Ma, Huang, Karpathy, Khosla, Bernstein, et~al.]{russakovsky2015imagenet}
Olga Russakovsky, Jia Deng, Hao Su, Jonathan Krause, Sanjeev Satheesh, Sean Ma, Zhiheng Huang, Andrej Karpathy, Aditya Khosla, Michael Bernstein, et~al.
\newblock Imagenet large scale visual recognition challenge.
\newblock \emph{International journal of computer vision}, 115:\penalty0 211--252, 2015.

\bibitem[Schmarje et~al.(2022)Schmarje, Grossmann, Zelenka, Dippel, Kiko, Oszust, Pastell, Stracke, Valros, Volkmann, et~al.]{schmarje2022one}
Lars Schmarje, Vasco Grossmann, Claudius Zelenka, Sabine Dippel, Rainer Kiko, Mariusz Oszust, Matti Pastell, Jenny Stracke, Anna Valros, Nina Volkmann, et~al.
\newblock Is one annotation enough?-a data-centric image classification benchmark for noisy and ambiguous label estimation.
\newblock \emph{Advances in Neural Information Processing Systems}, 35:\penalty0 33215--33232, 2022.

\bibitem[Shi et~al.(2023)Shi, Xu, Yuan, and Geng]{Shi2023Unreliable}
Yu Shi, Ning Xu, Hua Yuan, and Xin Geng.
\newblock Unreliable partial label learning with recursive separation.
\newblock In \emph{IJCAI}, pages 4208--4216, 2023.

\bibitem[Sohn et~al.(2020)Sohn, Berthelot, Carlini, Zhang, Zhang, Raffel, Cubuk, Kurakin, and Li]{sohn2020fixmatch}
Kihyuk Sohn, David Berthelot, Nicholas Carlini, Zizhao Zhang, Han Zhang, Colin~A Raffel, Ekin~Dogus Cubuk, Alexey Kurakin, and Chun-Liang Li.
\newblock Fixmatch: Simplifying semi-supervised learning with consistency and confidence.
\newblock \emph{Advances in neural information processing systems}, 33:\penalty0 596--608, 2020.

\bibitem[Szegedy et~al.(2016)Szegedy, Vanhoucke, Ioffe, Shlens, and Wojna]{szegedy2016rethinking}
Christian Szegedy, Vincent Vanhoucke, Sergey Ioffe, Jon Shlens, and Zbigniew Wojna.
\newblock Rethinking the inception architecture for computer vision.
\newblock In \emph{Proceedings of the IEEE conference on computer vision and pattern recognition}, pages 2818--2826, 2016.

\bibitem[Wang et~al.(2022)Wang, Xiao, Li, Feng, Niu, Chen, and Zhao]{wang2022pico}
Haobo Wang, Ruixuan Xiao, Yixuan Li, Lei Feng, Gang Niu, Gang Chen, and Junbo Zhao.
\newblock Pi{CO}: Contrastive label disambiguation for partial label learning.
\newblock In \emph{International Conference on Learning Representations}, 2022.

\bibitem[Wang et~al.(2023)Wang, Xiao, Li, Feng, Niu, Chen, and Zhao]{wang2022pico+}
Haobo Wang, Ruixuan Xiao, Yixuan Li, Lei Feng, Gang Niu, Gang Chen, and Junbo Zhao.
\newblock Pico+: Contrastive label disambiguation for robust partial label learning.
\newblock \emph{IEEE Transactions on Pattern Analysis and Machine Intelligence}, 2023.

\bibitem[Welinder et~al.(2010)Welinder, Branson, Mita, Wah, Schroff, Belongie, and Perona]{welinder2010caltech}
Peter Welinder, Steve Branson, Takeshi Mita, Catherine Wah, Florian Schroff, Serge Belongie, and Pietro Perona.
\newblock Caltech-ucsd birds 200.
\newblock 2010.

\bibitem[Wen et~al.(2021)Wen, Cui, Hang, Liu, Wang, and Lin]{wen2021leveraged}
Hongwei Wen, Jingyi Cui, Hanyuan Hang, Jiabin Liu, Yisen Wang, and Zhouchen Lin.
\newblock Leveraged weighted loss for partial label learning.
\newblock In \emph{International Conference on Machine Learning}, pages 11091--11100. PMLR, 2021.

\bibitem[Wu et~al.(2022)Wu, Wang, and Zhang]{wu2022revisiting}
Dong-Dong Wu, Deng-Bao Wang, and Min-Ling Zhang.
\newblock Revisiting consistency regularization for deep partial label learning.
\newblock In \emph{International Conference on Machine Learning}, pages 24212--24225. PMLR, 2022.

\bibitem[Xia et~al.(2023)Xia, Lv, Xu, Niu, and Geng]{xia2023towards}
Shiyu Xia, Jiaqi Lv, Ning Xu, Gang Niu, and Xin Geng.
\newblock Towards effective visual representations for partial-label learning.
\newblock In \emph{Proceedings of the IEEE/CVF Conference on Computer Vision and Pattern Recognition}, pages 15589--15598, 2023.

\bibitem[Xu et~al.(2024)Xu, Lian, Feng, Liu, and Tao]{xu2023alim}
Mingyu Xu, Zheng Lian, Lei Feng, Bin Liu, and Jianhua Tao.
\newblock Alim: Adjusting label importance mechanism for noisy partial label learning.
\newblock \emph{Advances in Neural Information Processing Systems}, 36, 2024.

\bibitem[Yao et~al.(2020)Yao, Deng, Chen, Gong, Wu, and Yang]{yao2020deep}
Yao Yao, Jiehui Deng, Xiuhua Chen, Chen Gong, Jianxin Wu, and Jian Yang.
\newblock Deep discriminative cnn with temporal ensembling for ambiguously-labeled image classification.
\newblock In \emph{Proceedings of the aaai conference on artificial intelligence}, pages 12669--12676, 2020.

\bibitem[Yu and Zhang(2016)]{yu2016maximum}
Fei Yu and Min-Ling Zhang.
\newblock Maximum margin partial label learning.
\newblock In \emph{Asian conference on machine learning}, pages 96--111. PMLR, 2016.

\bibitem[Zhang et~al.(2018)Zhang, Cisse, Dauphin, and Lopez-Paz]{zhang2017mixup}
Hongyi Zhang, Moustapha Cisse, Yann~N Dauphin, and David Lopez-Paz.
\newblock mixup: Beyond empirical risk minimization.
\newblock In \emph{International Conference on Learning Representations}, 2018.

\bibitem[Zhang and Yu(2015)]{zhang2015solving}
Min-Ling Zhang and Fei Yu.
\newblock Solving the partial label learning problem: An instance-based approach.
\newblock In \emph{IJCAI}, pages 4048--4054, 2015.

\bibitem[Zhang and Zhou(2013)]{zhang2013review}
Min-Ling Zhang and Zhi-Hua Zhou.
\newblock A review on multi-label learning algorithms.
\newblock \emph{IEEE transactions on knowledge and data engineering}, 26\penalty0 (8):\penalty0 1819--1837, 2013.

\end{thebibliography}
}


\end{document}